%% file: full-paper-template.tex
\documentclass[pmlr]{jmlr}

\RequirePackage{graphicx}
 \usepackage{booktabs}
\usepackage{longtable}
\makeatletter
\def\set@curr@file#1{\def\@curr@file{#1}} 
\makeatother
\usepackage[load-configurations=version-1]{siunitx} 

\usepackage{comment} 
\usepackage{multirow}
\usepackage[most]{tcolorbox}
\usepackage{soul}

\hypersetup{
  colorlinks=true,
  linkcolor=black,
  citecolor=blue,
  urlcolor=blue
}
 
\theorembodyfont{\upshape}
\theoremheaderfont{\scshape}
\theorempostheader{:}
\theoremsep{\newline}

\title[Diagnostic Uncertainty Preservation in Clinical Text]{Possible or Definite? A Benchmark for Evaluating Diagnostic Uncertainty Preservation in Clinical Text}

\author{\Name{Hongbo Du\nametag{\thanks{Equal contribution.}}}
       \Email{hdu24@my.trine.edu}\\ 
       \addr Trine University
      \AND
      \Name{Zixin Lu\nametag{\footnotemark[1]}}
      \Email{lzixin@umich.edu}\\ 
      \addr University of Michigan
      \AND
      \Name{Jiaming Qu\nametag{\thanks{Corresponding author. This research was conducted independently in a personal capacity and does not reflect the author's position at Amazon.}}}
      \Email{qjiaming@amazon.com}\\ 
      \addr Amazon}

\begin{document}

\maketitle

\input{tex/abstract}
\input{tex/introduction}
\input{tex/related_work}
\input{tex/method}
\input{tex/results}

\input{tex/discussion}

\input{tex/conclusion}

\bibliography{sample}

\newpage
\input{tex/appendix}

\end{document}

%% file: tex/abstract.tex
\begin{abstract}
Large language models (LLMs) are increasingly used for clinical text tasks such as summarization and revision. While most studies evaluate the fluency and coherence of LLM-generated text, whether LLMs correctly \emph{preserve diagnostic uncertainty} remains underexplored. In clinical practice, phrases such as ``possible pneumonia'' communicate the strength of available evidence and directly guide decisions about follow-up testing and treatment. Altering these uncertainty expressions can change the clinical meaning entirely. In this paper, we systematically evaluated this problem in two steps. First, we constructed a benchmark of 1,200 clinical documents with 9,184 uncertainty annotations across five levels. Second, we evaluated three LLMs on this benchmark. Our results show that (1) LLMs preserve the original uncertainty cues poorly, often less than half the time; (2) LLMs struggle with nuanced distinctions between adjacent levels. This work reveals a failure mode not captured by standard evaluation metrics and provides implications for the safe deployment of LLMs in clinical workflows.
\end{abstract}

%% file: tex/introduction.tex
\section{Introduction}
Large language models (LLMs) have been widely used in biomedical natural language processing (NLP) tasks such as clinical summarization, report generation, and question answering~\citep{tian2024opportunities}. Studies have shown that LLMs fine-tuned on specialized corpora can achieve human-level performance on complex clinical tasks~\citep{vanveen2024}. To systematically evaluate LLM performance on these tasks, prior work has introduced a variety of benchmarks. For example, ProbSum targets problem-list summarization~\citep{gao2023}, and \citep{xu2024} proposed a shared task for medical discharge note generation. However, existing evaluation metrics remain limited in scope, and many assessments do not directly measure real clinical impact~\citep{bednarczyk2025}.

Despite this progress, whether LLMs correctly interpret and preserve diagnostic uncertainty in clinical text remains underexplored. In clinical practice, phrases such as ``\underline{\emph{possible}} pneumonia'' and ``\underline{\emph{cannot exclude}} pulmonary embolism'' are not simply cautious wording. They communicate the strength of available evidence, indicate alternative diagnoses under consideration, and signal whether further testing is needed. These expressions directly influence clinical decision-making: a ``possible'' diagnosis may require additional imaging, while a definite assertion may lead to immediate treatment. Prior work has shown that uncertainty is a central and multidimensional component of clinical reasoning, and that clinicians use a wide range of uncertainty expressions in daily practice~\citep{han2011, mcgowan2025, panicekhricak2016}. However, LLMs can generate fluent and coherent text while altering the degree of uncertainty---for example, rewriting ``\emph{possible} pneumonia'' as ``pneumonia.'' If such changes propagate in clinical settings, they risk distorting the original clinical meaning. This issue has received little attention in prior work.

In this work, we study the following question: \textbf{Can LLMs correctly preserve diagnostic uncertainty in clinical NLP tasks?} Prior research on uncertainty in LLMs mainly follows two directions. The first studies uncertainty quantification, which uses logits, entropy, or related signals to detect unreliable outputs or hallucinations~\citep{farquhar2024}. The second studies uncertainty expressed \emph{within} LLM-generated text, evaluating whether models express uncertainty appropriately~\citep{kolagarzarcone2024, yang2025}. Our work follows the second direction. We do not measure the model's internal confidence during generation. Instead, we evaluate whether LLMs preserve the uncertainty that is already present in clinical source text on specific tasks.

To address this question, we constructed a benchmark for diagnostic uncertainty in clinical text. Our benchmark is based on two widely used datasets: MIMIC-IV-Note~\citep{johnson2023}, which contains discharge summaries and radiology reports, and TCGA-Reports~\citep{kefeli2024tcga}, a collection of pathology reports. We defined five levels of uncertainty, ranging from certain absent to non-asserted. To support fine-grained analysis, we used proposition-level annotations rather than assigning a single uncertainty label to each document. Each document is represented as a set of cue-target pairs, where each pair includes a target concept (e.g., ``pneumonia'') and its associated uncertainty cue (e.g., ``possible''). In total, our benchmark contains \textbf{1,200 documents across six clinical text types and 9,184 annotated cue-target pairs across five uncertainty levels}. We evaluated three LLMs on this benchmark using two complementary assessments: an indirect assessment that measures whether LLMs preserve uncertainty during text transformation tasks, and a direct assessment that tests whether an LLM can classify and rank uncertainty cues when explicitly prompted.

In summary, this paper systematically evaluates how LLMs handle diagnostic uncertainty in clinical NLP tasks. Our contributions are as follows. First, we constructed a benchmark covering multiple document types and uncertainty levels that can be reused in future work. Second, we provide an empirical evaluation of three LLMs across tasks and prompt conditions. Our results show that \textbf{LLMs frequently distort uncertainty: without appropriate prompting, they preserve the original uncertainty level less than half the time, and roughly two in five cue-target pairs are rewritten as definite assertions}. Additionally, LLMs classify individual uncertainty levels with moderate accuracy but struggle with fine-grained distinctions between adjacent levels. Together, these findings identify a systematic failure mode in LLM-based clinical text processing and highlight the urgent need for responsible LLMs usage in clinical workflows.


\subsection*{Generalizable Insights about Machine Learning in the Context of Healthcare}
\begin{itemize}
    \item Clinical meaning depends not only on \emph{which} condition is mentioned, but also on \emph{how certainly} it is expressed. Our results show that LLMs can produce fluent, factually grounded outputs that still distort meaning by altering uncertainty, suggesting that uncertainty preservation should be evaluated as a separate dimension.
    \item Evaluation results on our proposition-level benchmark reveals that the dominant failure mode is not random noise but a systematic bias toward certainty assertion---a pattern invisible to document-level evaluation. 
    \item Even explicit instructions to preserve uncertainty in the prompt are insufficient to fully prevent distortion, suggesting that more sophisticated interventions may be needed for safe and responsible LLM usage in clinical text processing tasks.
\end{itemize}

Our code and benchmark are publicly available at: \url{https://github.com/HongboD/Clinical-Uncertainty}

%% file: tex/related_work.tex
\section{Related Work}

\subsection{Uncertainty in Natural Language}
From a linguistic perspective, uncertainty is expressed through a wide range of forms rather than a single cue word. These include hedge words, modal verbs, differential diagnoses, incomplete evidence statements, and recommendations for follow-up testing~\citep{Kilicoglu2008SpeculativeLanguage}. These expressions are particularly important in biomedical settings, where they communicate limits of evidence, possible alternatives, and planned next steps in patient care. Prior work has shown that uncertainty in healthcare is multidimensional rather than incidental, spanning a wide range of linguistic forms and clinical functions~\citep{han2011}. Clinical NLP studies have developed resources for detecting related phenomena such as negation, hedging, and assertion status~\citep{uzuner2011, vincze2008, peng2018}. However, these datasets are limited in the number of uncertainty levels, the types of clinical text, and the overall scale. We extend this line of work by constructing a benchmark with broader coverage of uncertainty levels, document types, and dataset size.

\subsection{Evaluating LLMs' Awareness of Uncertainty}
Prior work on studying uncertainty in LLMs can be grouped into two main directions. The first direction studies LLMs' internal uncertainty, including methods based on logits, entropy, calibration, and related signals that estimate the likelihood of errors or hallucinations~\citep{farquhar2024, kadavath2022language, kapoor2024calibration}. This line of work evaluates how reliably LLMs generate responses based on their internal state. The second direction studies uncertainty expressed in LLM-generated language, examining whether uncertainty is expressed appropriately in the output. Examples include uncertainty transfer in summarization and long-form text generation~\citep{kolagarzarcone2024, yang2025, yang2025uncle}, as well as uncertainty-aware diagnosis and explanation~\citep{zhou2025uncertainty}. Our work follows the second direction. Specifically, we focus on patient-specific clinical notes and evaluate whether diagnostic uncertainty is preserved from source text to output.

\subsection{Benchmarks in Clinical LLMs}
Prior studies have curated various benchmarks to evaluate LLMs across different tasks, including clinical note summarization, report generation, and question answering~\citep{gao2023drbench,vanveen2024,xu2024,kweon2024ehrnoteqa,liu2024clinicbench}. These benchmarks mainly focus on evaluating the fluency and coherence of LLM-generated text and whether LLMs can produce accurate responses. However, a plausible and coherent output does not guarantee clinically meaningful assessment~\citep{bednarczyk2025,bedi2025testing,gong2025knowledgegap}. To address this gap, we constructed a benchmark to evaluate LLMs' understanding of diagnostic uncertainty. Motivated by prior work, our benchmark construction combines two approaches: (1) deriving supervision from existing clinical document collections~\citep{gao2023} and (2) combining automated data extraction with human review~\citep{kweon2024ehrnoteqa,grazhdanski2025synthmedic}.

%% file: tex/method.tex
\section{Method}

\begin{figure}[hbtp!]
  \centering
  \includegraphics[width=1\linewidth]{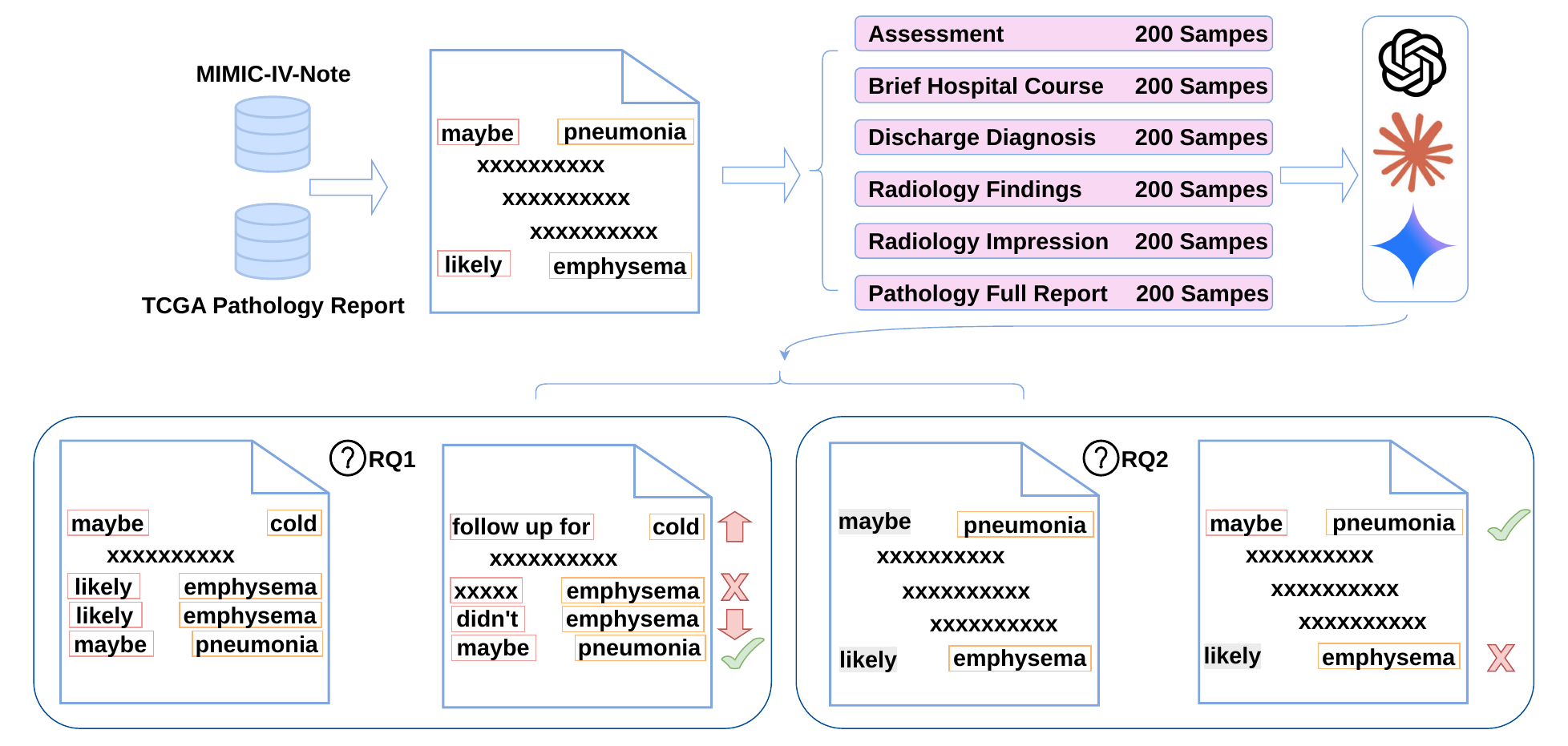}
  \caption{Overview of the study workflow.}
  \label{fig:overview}
\end{figure}

The main goal of this study is to evaluate \textbf{whether LLMs can correctly interpret uncertainty cues in clinical text}. We do not evaluate LLMs on traditional measures such as generation quality, coherence, or factual accuracy. Instead, we study two complementary research questions (RQs):

\begin{itemize}
  \item \textbf{RQ1 (Indirect assessment):} Do LLMs preserve diagnostic uncertainty when transforming clinical text in downstream tasks such as summarization and patient-friendly rewriting?
  \item \textbf{RQ2 (Direct assessment):} Can LLMs correctly identify and rank uncertainty cues when explicitly asked to interpret them?
\end{itemize}

To study these RQs, we took two steps (Figure~\ref{fig:overview}). First, because no suitable dataset was readily available, we constructed a benchmark of 1,200 clinical documents spanning six clinical text types. We defined five uncertainty levels, ranging from certain absent to non-asserted. Second, we evaluated three LLMs on this benchmark. We describe our benchmark construction (Section~\ref{subsec:benchmark_construction}) and evaluation approaches (Section~\ref{subsec:eval}) below.

\subsection{Benchmark Construction}\label{subsec:benchmark_construction}

\textbf{Data Pre-processing.} We constructed the benchmark from two datasets that are widely used in biomedical NLP research: (1) MIMIC-IV-Note~\citep{johnson2023}, which contains 2.6M clinical notes, and (2) TCGA-Reports~\citep{kefeli2024tcga}, a dataset of 9,523 pathology reports. For MIMIC-IV-Note, we focused on two document types: discharge summaries and radiology reports, as they are relatively long and provide rich material for studying uncertainty. Each MIMIC note is organized into sections, and we selected the most clinically relevant ones: assessment, brief hospital course, and discharge diagnosis for discharge summaries; impressions and findings for radiology reports. Each section was extracted as a standalone document. For TCGA pathology reports, we used the full document because structured sections were not available. In total, our initial document pool included six document types, all of which are interpretive clinical text in which clinicians record conclusions, exclusions, contingencies, and unresolved issues about the patient.

\textbf{Defining Uncertainty Cues.} A key step in our benchmark construction was extracting uncertainty cue-target pairs from each document. We define an \textit{uncertainty cue} as a text span (a word or phrase) that expresses the writer's degree of commitment toward a clinical proposition. The \textit{target} is the clinical proposition modified by that cue. For example, in the sentence \textit{``Limited evaluation due to suboptimal contrast timing; no large central filling defect is seen, but \underline{pulmonary embolism} \underline{cannot be excluded},''} ``cannot be excluded'' is the uncertainty cue and ``pulmonary embolism'' is the target.

Informed by prior work on clinical uncertainty, negation, and assertion status in biomedical text~\citep{han2011,Bhise2018,mcgowan2025,vincze2008,uzuner2011,peng2018,limalopez2020,thompson2011,Callen2020}, we defined a five-level uncertainty schema. The five levels range from certain absent to non-asserted mentions:

\begin{itemize}
    \item \textbf{Certain absent:} The clinical proposition is not present for the patient, expressed with clear and confident language that leaves no doubt about its absence.
    \item \textbf{Probable present:} The clinical proposition is more likely present than absent, but the language indicates that the writer is not fully certain.
    \item \textbf{Possible present:} The clinical proposition may be present, but the language indicates that this is only one reasonable possibility rather than a confident conclusion.
    \item \textbf{Indeterminate or unresolved:} The clinical proposition cannot yet be determined because the available information is incomplete, mixed, or unclear.
    \item \textbf{Non-asserted evaluation target:} The clinical proposition is mentioned as something to check, test for, monitor, or rule out. The text does not state that it is present or absent; it is only the target of evaluation.
\end{itemize}

We began with an uncertainty cue inventory drawn from the studies above, identifying approximately five common cues for each level. To increase coverage, we took two approaches. First, each author independently reviewed a small set of MIMIC notes and conducted qualitative coding of sentences expressing uncertainty. All authors then met twice to discuss and select the most frequent cues. Second, we invited a linguist and a clinician to a discussion panel in which all authors participated for final refinement. This step helped capture conventions specific to clinical notes, such as the use of the question mark as a common symbol for non-asserted information. Table~\ref{tab:uncertainty_levels} summarizes definitions and representative cues for each level. UL3 (\emph{Possible present}) has the most unique cues ($N=30$), while UL1 (\emph{Certain absent}) has the fewest ($N=9$). A full list of cues for each level is provided in Appendix~\ref{appendix:full_list_of_cues}.

\begin{table}[t]
  \centering
  \caption{We defined five uncertainty levels (ULs), with UL5 representing the most uncertainty and UL1 representing the least. We developed a set of cue words for each level. $N$ denotes the unique number of cues for each level.}
  \small
  \begin{tabular}{lp{2.8cm}p{8.6cm}}
    \toprule
    \textbf{Level} & \textbf{Type} & \textbf{Uncertainty cues} \\
    \midrule
    UL1 & Certain absent & ($N=9$): no; without; denies; negative for; absent; free of; no evidence of; ruled out \\
    UL2 & Probable present & ($N=21$): likely; probable; suggests; suspicious for; concerning for; favored; appears to represent \\
    UL3 & Possible present & ($N=30$): possible; may; might; could; cannot exclude; may represent; raises the possibility of; question of \\
    UL4 & Indeterminate & ($N=17$): unclear; unresolved; indeterminate; equivocal; uncertain; cannot determine; difficult to assess \\
    UL5 & Non-asserted & ($N=14$): ?, rule out; follow up for; follow-up to exclude; workup for; query; concern for; contingency \\
    \bottomrule
  \end{tabular}
  \label{tab:uncertainty_levels}
\end{table}

\textbf{Cue-Target Pair Extraction.} To extract cue-target pairs from each document, we built a rule-based pipeline that operates in two stages: extraction and refinement. We deliberately chose a rule-based approach rather than using LLMs to avoid circular evaluation, where the same type of model would both generate the benchmark labels and be evaluated against them. We also prioritized precision over recall, as the benchmark requires reliable labels rather than exhaustive coverage. Our pipeline proceeds as follows:

\begin{itemize}
    \item \textbf{Step 1: Extraction.} We first segment each document into sentences and scan for uncertainty cues using regular expressions (regex) over the cue inventory defined in Table~\ref{tab:uncertainty_levels}. Each cue pattern encodes a syntactic direction: some cues modify the phrase that follows (e.g., ``\textit{no evidence of} pneumonia''), while others modify the phrase that precedes them (e.g., ``the adrenal nodule \textit{is indeterminate}''). When a cue is detected, we extract a candidate target span from the appropriate direction within the same sentence. If the candidate span contains coordination structures (e.g., ``PE or pneumonia'' after ``negative for''), we split it into separate targets and create one cue-target pair per target. 
    \item \textbf{Step 2: Refinement.} The extraction stage produces candidate pairs whose target spans are raw text fragments. These fragments may include extraneous tokens (e.g., trailing punctuation, leading articles, overly broad clauses) or may miss the precise medical concept boundary. The refinement stage addresses this by independently extracting medical concepts from the same text and linking them back to the pairs from Step 1. This stage consists of three consecutive sub-steps.
    \item \textbf{Step 2a: Medical Concept Extraction.} For each sentence, we run medSpaCy~\citep{eyre2022launching} for medical named entity recognition. We then use QuickUMLS~\citep{soldaini2016quickumls} to match each entity to a Concept Unique Identifier (CUI) in the Unified Medical Language System~\citep{bodenreider2004unified}. This step produces a set of medical concept candidates for each sentence.
    \item \textbf{Step 2b: Candidate Scoring and Linking.} For each cue-target pair from Step 1, we score every medical concept candidate from Step 2a based on span overlap with the original regex capture, character proximity to the cue, and directional consistency with the cue's syntactic pattern. If all candidates score below a minimum threshold, the pair is flagged for manual review.
    \item \textbf{Step 2c: Target Replacement.} When a high-confidence medical concept is found, we replace the raw regex-captured target span with the matched medical concept. If no suitable match exists for a given target---which is common for clinical abbreviations (e.g., ``FNH,'' ``PNA,'' ``AMS'')---we retain the original target from Step 1.
\end{itemize}

After applying the pipeline to each sentence within a document, we perform deduplication on the target text to obtain a set of unique targets. For target texts with exact string matches, we prioritize the pair that fills a gap in the document's uncertainty level coverage. For target texts that differ lexically but share the same CUI, we prioritize the more specific surface form. Finally, each document has a set of cue-target pairs with unique targets.

\textbf{Sampling.} After extracting cue-target pairs from each document, we conducted a pre-screening step by excluding documents that covered fewer than two unique uncertainty levels. These documents provided insufficient uncertainty-related content for the benchmark. We then sampled from the pre-screened documents, giving higher priority to documents with broader coverage of uncertainty levels. For example, documents containing pairs from all five levels were sampled first. We sampled 200 documents from each of the six document types, resulting in a final benchmark of 1,200 documents.

We performed human validation for quality control. We randomly sampled 15 documents from each document type, for a total of 90 documents (7.5\% of the benchmark). Two authors with medical backgrounds independently reviewed all extracted target texts and their associated cues, covering 891 pairs in total. For each pair, they annotated whether \emph{both} the target text and the cue were extracted correctly. The results showed an average accuracy of 82\%, with an inter-reviewer agreement of Cohen's $\kappa = 0.61$, indicating moderate agreement~\citep{landis1977measurement}. These results suggest that the rule-based pipeline achieved reasonable quality for benchmark construction.

\textbf{Benchmark Statistics.} The final benchmark contains 1,200 documents across six clinical document types, with 200 documents per type: assessment, brief hospital course, discharge diagnosis, radiology findings, radiology impression, and full pathology reports. Document length ranges from 17 to 4,013 words (mean = 572.41; S.D.\ = 482.4). Because we prioritized documents with broader uncertainty level coverage during sampling, each document covers 3.54 unique uncertainty levels on average. There is slight variance across document types due to the nature of the data: brief hospital course documents tend to cover all five levels because they are longer and describe key events, treatments, and outcomes during a patient's stay; in contrast, assessment, discharge diagnosis, and TCGA reports are dominated by documents covering only three levels, as they tend to be conclusion-oriented rather than focused on clinical findings. In total, the benchmark contains 9,184 cue-target pairs,\footnote{The detailed distribution across uncertainty levels is: 3,233 pairs in UL1, 2,147 in UL2, 2,631 in UL3, 467 in UL4, and 706 in UL5.} with an average of 7.65 pairs per document.

\subsection{Evaluations}\label{subsec:eval}

Our RQ1 evaluates whether LLMs preserve uncertainty when performing downstream text transformation tasks. Our RQ2 evaluates whether LLMs can directly identify and rank uncertainty cues when explicitly prompted. We describe the experimental setup and metrics for each RQ below.

\subsubsection{RQ1: Indirect Assessment}\label{subsec:rq1}
\textbf{Experimental design.} We evaluate LLMs on two free-text clinical transformation tasks. The first task is \textit{clinician handoff summarization}, in which the model produces a concise clinician-facing summary for continuity of care. This task places strong pressure on compression and reorganization, making it a natural setting for observing uncertainty preservation or erosion. The second task is \textit{patient-friendly revision}, in which the model rewrites the source text in plain language while preserving its meaning.

Each task is evaluated under two prompt conditions. In the \textsc{Baseline} condition, the model receives only the core task instruction and the source text. In the \textsc{Guarded} condition, the instruction additionally requires the model to rely only on source information, avoid adding new diagnoses or interpretations, and preserve uncertain or hedged statements rather than converting them into definite claims. This yields a 2 (task) $\times$ 2 (prompt condition) design. All prompts are provided in Appendix~\ref{appendix:rq1_prompt}.

\textbf{Models and infrastructure.} We evaluated three LLMs: \texttt{gpt-oss-120b}~\citep{OpenAI_gptoss120b_2025}, \texttt{gemini-2.5-flash}~\citep{Google_Gemini25Flash_2025}, and \texttt{claude-haiku-4.5}~\citep{Anthropic_ClaudeHaiku45_2025}. To comply with data retention policies for the MIMIC-IV-Note dataset, we ran the three models on separate platforms: Groq\footnote{\url{https://console.groq.com/docs}} for \texttt{gpt-oss-120b}, Google Vertex AI\footnote{\url{https://cloud.google.com/vertex-ai}} for \texttt{gemini-2.5-flash}, and AWS Bedrock\footnote{\url{https://aws.amazon.com/bedrock/}} for \texttt{claude-haiku-4.5}. The decoding temperature was set to 0 for all models, as we focus on faithful transformation rather than response diversity.

\textbf{Metrics.} To evaluate RQ1, we need to determine whether each source cue-target pair is preserved in the LLM-generated text. Let $\mathcal{S}$ denote the set of all cue-target pairs in the source document. For each pair, we search the LLM output for the target using exact string match, normalized lexical forms, or phrases linked to the same CUI. This produces two subsets: $\mathcal{S}_+$, the pairs whose targets are retained in the output, and $\mathcal{S}_-$, the pairs whose targets are absent. We define five metrics:

\begin{itemize}
    \item \textit{Target Retained Rate (TRR):} The proportion of source pairs whose targets appear in the model output (i.e., $|\mathcal{S}_+| / |\mathcal{S}|$). TRR measures whether the model keeps the clinical proposition at all, regardless of its uncertainty level.
    \item \textit{Uncertainty Retained Rate (URR):} Among $\mathcal{S}_+$, the proportion of pairs for which the model preserves the same uncertainty level as the source. URR is the primary metric, as it directly measures whether the model faithfully reproduces the original degree of uncertainty.
    \item \textit{Certainty Assertion Rate (CAR):} Among $\mathcal{S}_+$, the proportion of pairs for which the model removes the uncertainty cue entirely, rewriting the target as a definite assertion. CAR captures the most clinically concerning form of distortion: collapsing hedged language into certain claims.
    \item \textit{Partial Certainty Rate (PCR):} Among $\mathcal{S}_+$, the proportion of pairs for which the model shifts the uncertainty level toward certainty without reaching full assertion. PCR complements CAR by capturing partial movement toward certainty.
    \item \textit{Over-hedging Rate (OHR):} Among $\mathcal{S}_+$, the proportion of pairs for which the model shifts the uncertainty level away from certainty. OHR captures cases where the model introduces more hedging than the original text.
\end{itemize}

\subsubsection{RQ2: Direct Assessment}\label{subsec:rq2}

\textbf{Experimental design.} We evaluate whether LLMs can directly interpret clinical uncertainty when explicitly prompted. We sampled 500 cue-target pairs from the benchmark, with 100 pairs from each uncertainty level. Each sample contains a short sentence snippet, a target concept, and a ground truth uncertainty level. The model is prompted to (1) identify the shortest quote in the snippet that signals the uncertainty status of the target, and (2) classify the corresponding uncertainty level. We provide the definitions of UL1--UL5 in the prompt. We use \texttt{gemini-2.5-flash} for this experiment because it was the strongest-performing model in RQ1.

We evaluate two settings. In the single-signal setting ($N{=}1$), the model receives one cue-target pair at a time and classifies its uncertainty level. In the multi-signal setting ($N{=}2, 3, 4, 5$), the model receives a bundle of $N$ cue-target pairs simultaneously and must classify each pair and rank them from strongest to weakest certainty. The multi-signal setting tests whether the model can distinguish between uncertainty levels when several signals must be compared jointly. For the multi-signal setting, we construct 100 bundles for each value of $N$, yielding 400 bundles in total across $N{=}2$--$5$. The full prompts for both settings are provided in Appendix~\ref{appendix:rq2_prompt}.

\textbf{Metrics.} We define four metrics for RQ2:

\begin{itemize}
    \item \textit{Uncertainty Level Accuracy (ULA):} The proportion of pairs for which the predicted uncertainty level exactly matches the gold label. ULA is the primary classification metric and is reported for all settings.

    \item \textit{Macro-F1:} The unweighted mean of per-level F1 scores across UL1--UL5. Each level contributes equally regardless of classification difficulty. Macro-F1 complements ULA by accounting for potential imbalances in per-level performance.

    \item \textit{Per-level F1:} The F1 score computed separately for each uncertainty level. Per-level F1 reveals which levels the model handles well and which it struggles with.

    \item \textit{Exact Rank Accuracy:} The proportion of bundles for which the model predicts the entire strongest-to-weakest ordering correctly. This is only for the $N{=}2$--$5$ setting. 
\end{itemize}

%% file: tex/results.tex
\section{Results}
\label{sec:results}

\subsection{RQ1: Indirect Assessment}
\label{subsec:rq1_results}

Table~\ref{tab:rq1_results} reports all five metrics for the three LLMs across both tasks and prompt conditions. We identify five trends.

\begin{table*}[htbp!]
  \centering
  \caption{RQ1 results. All values are percentages. Higher TRR and URR indicate better preservation; higher CAR, PCR, and OHR indicate more distortion.}
  \small
  \begin{tabular}{llccccc}
    \toprule
    \textbf{Model} & \textbf{Task / Condition} & \textbf{TRR} & \textbf{URR} & \textbf{CAR} & \textbf{PCR} & \textbf{OHR} \\
    \midrule
    \multirow{4}{*}{\texttt{claude-haiku-4.5}}
      & Revision / Baseline       & 32.82 & 33.64 & 44.96 & 3.88 & 1.26 \\
      & Revision / Guarded        & 36.57 & 43.35 & 40.07 & 3.57 & 1.37 \\
      & Summarization / Baseline  & 55.25 & 43.32 & 41.90 & 5.30 & 1.73 \\
      & Summarization / Guarded   & 67.79 & 62.51 & 24.59 & 4.22 & 1.78 \\
    \midrule
    \multirow{4}{*}{\texttt{gemini-2.5-flash}}
      & Revision / Baseline       & 46.54 & 38.98 & 42.65 & 4.16 & 1.54 \\
      & Revision / Guarded        & 48.41 & 46.06 & 37.04 & 3.89 & 1.50 \\
      & Summarization / Baseline  & 49.99 & 45.55 & 38.12 & 4.98 & 2.24 \\
      & Summarization / Guarded   & 62.45 & 62.53 & 23.64 & 3.54 & 1.95 \\
    \midrule
    \multirow{4}{*}{\texttt{gpt-oss-120b}}
      & Revision / Baseline       & 48.08 & 37.77 & 42.80 & 3.79 & 2.01 \\
      & Revision / Guarded        & 50.59 & 47.72 & 34.42 & 4.10 & 1.63 \\
      & Summarization / Baseline  & 53.44 & 44.36 & 37.88 & 6.12 & 1.83 \\
      & Summarization / Guarded   & 59.21 & 56.18 & 29.28 & 5.36 & 1.69 \\
    \bottomrule
  \end{tabular}
  \label{tab:rq1_results}
\end{table*}

\textbf{Models frequently omit clinical targets.} Under the \textsc{Baseline} condition, TRR ranges from 32.82\% to 55.25\%, indicating that LLMs often drop clinical targets entirely. Revision yields substantially lower TRR than summarization across all models. For example, \texttt{claude-haiku-4.5} retains only 32.82\% of targets during revision compared to 55.25\% during summarization. This suggests that patient-friendly rewriting leads to more aggressive content removal than clinician handoff summarization.

\textbf{Uncertainty preservation is poor under baseline.} Under the \textsc{Baseline} condition, URR ranges from 33.64\% to 45.55\% across all model-task combinations, indicating that even when a model retains a clinical target, it preserves the original uncertainty level correctly less than half the time. The pattern is consistent across all three models. Summarization yields slightly higher URR than revision (e.g., 45.55\% vs.\ 38.98\% for \texttt{gemini-2.5-flash}), suggesting that the more extensive rewriting required by revision makes uncertainty more vulnerable to distortion.

\textbf{Distortion is overwhelmingly toward certainty.} Among the three distortion metrics (CAR, PCR, OHR), certainty assertion is by far the dominant failure mode. Under the \textsc{Baseline} condition, CAR ranges from 37.88\% to 44.96\%, meaning that roughly two in five retained targets are rewritten as definite assertions with the uncertainty cue removed entirely. In contrast, PCR ranges from 3.79\% to 6.12\% and OHR ranges from 1.26\% to 2.24\%. Together, partial certainty shifts and over-hedging account for less than 8\% of retained pairs. This indicates that when LLMs distort uncertainty, they almost always collapse hedged language into definite assertions rather than make subtle shifts along the uncertainty scale.

\textbf{Guarded prompting helps but does not eliminate distortion.} Comparing the \textsc{Guarded} condition to the \textsc{Baseline}, URR increases for all six model-task combinations, with gains ranging from +7.08 to +19.19 percentage points. The improvements are consistently larger for summarization than for revision. CAR also decreases under the \textsc{Guarded} condition by 4.89--17.31 percentage points. However, even the best guarded configuration achieves only 62.53\% URR, leaving substantial room for improvement.

\textbf{Variation across tasks.} Across all models and prompt conditions, summarization consistently outperforms revision on both TRR and URR. This pattern likely reflects the nature of the two tasks: summarization compresses content for a clinical audience that expects hedged language, while revision rewrites for patients and tends to simplify or remove clinical nuance. The gap is most pronounced for \texttt{claude-haiku-4.5}, where baseline summarization URR exceeds baseline revision URR by nearly 10 percentage points.

\subsection{RQ2: Direct Assessment}
\label{subsec:rq2_results}

Table~\ref{tab:rq2_by_n} reports the overall classification and ranking metrics across the single-signal and multi-signal settings. Table~\ref{tab:rq2_per_level} reports per-level F1 scores. We only experimented with \texttt{gemini-2.5-flash} based on the overall best performance on RQ1. We identify three trends.

\textbf{Classification is moderately accurate; ranking degrades with signal load.} In the single-signal setting ($N{=}1$), it achieves 78.4\% ULA and 78.6\% Macro-F1 (Table~\ref{tab:rq2_by_n}). When the model receives multiple signals simultaneously ($N{=}2$--$5$), classification accuracy remains stable, indicating that additional signals do not impair per-item labeling. However, exact rank accuracy decreases steadily from 86.0\% at $N{=}2$ to 48.0\% at $N{=}5$, as the number of possible orderings grows factorially. Kendall's $\tau$ remains above 0.72 across all values of $N$, confirming that predicted rankings stay positively correlated with the gold rankings even when the exact order is incorrect.

\begin{table}[htbp!]
  \centering
  \caption{RQ2 results by number of simultaneously presented signals ($N$).}
  \small
  \begin{tabular}{cccc}
    \toprule
    $N$ & \textbf{ULA} & \textbf{Macro-F1} & \textbf{Exact Rank Accuracy} \\
    \midrule
    1 & 78.4 & 78.6 & --- \\
    2 & 80.5 & 79.8 & 86.0 \\
    3 & 80.3 & 80.0 & 78.0 \\
    4 & 82.0 & 81.8 & 68.0 \\
    5 & 81.6 & 81.3 & 48.0 \\
    \midrule
    2--5 & 81.3 & 81.0 & 70.0 \\
    \bottomrule
  \end{tabular}
  \label{tab:rq2_by_n}
\end{table}

\textbf{Per-level performance and the effect of contrastive signals.} In the single-signal setting, the model performs best on UL1 (Certain absent), which involves clear negation cues, and worst on UL3 (Possible present), a level that requires fine-grained distinction from adjacent levels (Table~\ref{tab:rq2_per_level}). This pattern suggests that the model handles categorical distinctions (present vs.\ absent) more reliably than graded ones along the uncertainty spectrum. In the multi-signal setting, the F1 scores for UL2 and UL3 increases by 10.5 and 9.2 points respectively compared to $N{=}1$, suggesting that contrastive signals from other levels help the model calibrate the boundary between probable and possible. In contrast, F1 for UL4 and UL5 decreases by 3.2 and 4.8 points, indicating that indeterminate and non-asserted levels become harder to distinguish when multiple signals are presented simultaneously.

\begin{table}[htbp!]
  \centering
  \caption{RQ2 per-level F1 scores under the single-signal ($N{=}1$) and multi-signal ($N{=}2$--$5$, averaged) settings.}
  \small
  \begin{tabular}{lcc}
    \toprule
    \textbf{Level} & $N{=}1$ & $N{=}2$--$5$ \\
    \midrule
    UL1 (Certain absent)    & 94.6 & 94.8 \\
    UL2 (Probable present)  & 72.2 & 82.7 \\
    UL3 (Possible present)  & 67.0 & 76.2 \\
    UL4 (Indeterminate)     & 83.1 & 79.9 \\
    UL5 (Non-asserted)      & 76.1 & 71.3 \\
    \bottomrule
  \end{tabular}
  \label{tab:rq2_per_level}
\end{table}

%% file: tex/discussion.tex
\section{Discussion}
\subsection{Summary of Findings}
Our study evaluated whether LLMs correctly preserve diagnostic uncertainty in clinical text through two complementary assessments on a benchmark of 1,200 documents with 9,184 proposition-level uncertainty annotations across five levels. 

In the indirect assessment (RQ1), we found that LLMs frequently distort uncertainty during both summarization and revision. Under baseline prompting, all three LLMs preserve the original uncertainty level less than half the time. The dominant failure mode is certainty assertion: roughly two in five retained propositions are rewritten as definite claims with the uncertainty cue removed entirely. LLMs also frequently omit clinical targets altogether, particularly during revision, where the more extensive rewriting may cause models to drop uncertainty-bearing content. The \textsc{Guarded} condition slightly improves uncertainty preservation across all models and tasks.

In the direct assessment (RQ2), \texttt{gemini-2.5-flash} classifies uncertainty levels with moderate accuracy when presented with a single signal. Classification accuracy remains stable as the number of simultaneously presented signals increases, but exact ranking accuracy degrades from 86\% at $N{=}2$ to 48\% at $N{=}5$. Per-level analysis reveals that the model handles categorical distinctions reliably but struggles with fine-grained distinctions along the uncertainty spectrum, particularly between probable and possible. 

\subsection{Implications for Using LLMs in Clinical Tasks}
Our study offers three implications for using LLMs in clinical tasks.

First, our findings suggest that uncertainty preservation should be treated as a distinct evaluation dimension for clinical LLM systems. Current evaluation of clinical text generation focuses on content coverage, factual consistency, and fluency, yet recent reviews have noted that safety-relevant evaluation remains underdeveloped~\citep{bednarczyk2025,bedi2025testing}. We computed BERTScore for LLM outputs in RQ1 (Appendix~\ref{appendix:bertscore}), and all conditions achieved scores around 82\%. However, despite appearing fluent and factually plausible, the LLM-generated outputs still changed the clinical meaning by altering the degree of certainty attached to a diagnosis. This can lead to harmful consequences in healthcare, where uncertainty communicates the strength of evidence, the openness of the differential diagnosis, and the need for further testing~\citep{han2011,Bhise2018}. Our study highlights the importance of developing benchmarks and metrics that evaluate uncertainty preservation to support safe and responsible use of LLMs in clinical tasks.

Second, the certainty distortion is systematic rather than random. Across models and tasks, the dominant error mode is movement toward certainty by removing uncertainty cues entirely, e.g., \textit{``\st{possible} pneumonia''}. This is especially concerning because the most dangerous distortion is not adding false information, but converting a tentative proposition into a definite one. One plausible explanation is that LLMs inherit a bias from training data in which hedges are often removed for simplicity~\citep{kolagarzarcone2024}. This interpretation is reinforced by our RQ2 results: the model can classify uncertainty with moderate accuracy when explicitly asked, yet fails to preserve those same cues during downstream tasks. This gap suggests that the bottleneck is not whether the model can recognize uncertainty, but whether it prioritizes preserving uncertainty while optimizing for compression and fluency~\citep{gong2025knowledgegap}. 

Finally, our results show that guarded prompting helps but is not sufficient, and that vulnerability varies by task. The \textsc{Guarded} condition improved URR by 7--19 percentage points, but even the best configuration reached only about 63\% URR. Moreover, patient-friendly revision is consistently more vulnerable than clinician-facing summarization across all models and prompt conditions. This matters because patient-facing text is where uncertainty distortion has the most direct impact: patients and caregivers typically cannot independently assess the underlying evidence and may take a definite-sounding statement at face value. Together, these findings suggest that prompt engineering should be viewed as a partial mitigation rather than a complete solution. Future work can explore stronger interventions such as fine-tuning on our benchmark, preference optimization with uncertainty-aware reward signals, or post-hoc verification~\citep{yang2025,yang2025uncle,zhou2025uncertainty}.

\subsection{Limitations and Future Work}
Our study has several limitations. First, the benchmark was constructed using a rule-based pipeline rather than full manual annotation. Although human validation showed reasonable quality (82\% accuracy, Cohen's $\kappa = 0.61$), some extraction errors may remain. Second, we evaluated three general-purpose LLMs as off-the-shelf tools. While the \textsc{Guarded} condition suggests improvement over the baseline, we did not fine-tune on clinical data or incorporate uncertainty-aware reward signals, which may further improve performance. Finally, perceptions of uncertainty are inherently subjective and may vary across clinicians. We plan to investigate whether distorted uncertainty leads to different clinical decisions through clinician-facing user studies that measure real downstream impact. We also plan to explore a multilingual uncertainty cue inventory in future work.

%% file: tex/conclusion.tex
\section{Conclusion}
In this paper, we presented a systematic evaluation of whether LLMs preserve diagnostic uncertainty in clinical text. We constructed a benchmark of 1,200 clinical documents with 9,184 proposition-level uncertainty annotations across five levels and evaluated three LLMs using two complementary assessments. Our results show that LLMs frequently distort uncertainty during text transformation, with the dominant failure mode being the collapse of hedged language into definite assertions. Guarded prompting reduces but does not eliminate this distortion. When directly prompted to classify and rank uncertainty levels, the model classifies individual levels with moderate accuracy but struggles with fine-grained distinctions between adjacent levels. These findings identify a clinically important failure mode not captured by standard evaluation metrics. We will release our benchmark to support future work and encourage joint effort between computer scientists and clinicians for safe and responsible usage of LLMs in clinical NLP tasks.

%% file: tex/appendix.tex
\appendix

\section{Full List of Uncertainty Cues}\label{appendix:full_list_of_cues}

All uncertainty cues used in this study are summarized below:
\begin{itemize}
    \item UL1 ($N=9$): no; no evidence of; negative for; without evidence of; free of; denies; denied; absence of; absent
    \item UL2 ($N=21$): likely; consistent with; most likely; suggestive of; probable; likely due to; probably; compatible with; appears to be; likely from; likely represents; likely secondary to; most consistent with; likely represent; favored; most likely represents; favors; most likely represent; favored to represent; seems to be; favoured to represent
    \item UL3 ($N=31$): possible; concern for; concerning for; possibly; may be; could be; may represent; suspicious for; cannot be excluded; suspected; could represent; suspect; may reflect; question of; questionable; differential includes; could reflect; cannot be fully excluded; differential diagnosis includes; cannot rule out; difficult to exclude; might be; might represent; is not excluded; cannot exclude; cannot be completely excluded; might reflect; are not excluded; can't rule out; cannot be entirely excluded 
    \item UL4 ($N=17$): unclear; indeterminate; too small to characterize; not well visualized; equivocal; not clearly; cannot be determined; too small to fully characterize; not clear; not entirely clear; nondiagnostic; degraded by motion; limited evaluation for; suboptimal for; motion degraded; unknown; cannot determine
    \item UL5 ($N=14$): ?; rule out; to exclude; evaluate for; r/o; monitor for; workup for; query; watch for; follow up to exclude; possible; could be; follow-up to exclude; repeat ct 
\end{itemize}

\section{Prompts in RQ1 Experiments}\label{appendix:rq1_prompt}
\begin{tcolorbox}[
    colback=black!5,
    colframe=black!40,
    arc=2mm,
    boxrule=0.8pt,
    left=2mm,
    right=2mm,
    top=2mm,
    bottom=2mm,
    title=\textbf{Baseline Prompt for the Summarization task}
]
\small
You are a clinical handoff assistant helping prepare a concise clinician-to-clinician summary for the next provider taking over care.

Your task is to summarize the source text into a brief clinician-facing handoff summary.

The output will be read by another clinician during a care transition. Focus on the most clinically important information needed for continuity of care, including major problems, key findings, current status, and important near-term considerations. Keep the summary concise, accurate, and professionally written.

\{\{input text\}\}
\end{tcolorbox}

\begin{tcolorbox}[
    colback=black!5,
    colframe=black!40,
    arc=2mm,
    boxrule=0.8pt,
    left=2mm,
    right=2mm,
    top=2mm,
    bottom=2mm,
    title=\textbf{Baseline Prompt for the Revision task}
]
\small
You are a patient education assistant helping rewrite clinical text into plain-language information that a patient or family member can understand.

Your task is to rewrite the source text in clear, patient-friendly language without changing its meaning.

The output will be read by a patient or caregiver. Use plain, direct, non-technical language where possible. Preserve the clinical meaning, but make the wording easier to understand. Keep the tone respectful, calm, and supportive.

\{\{input text\}\}
\end{tcolorbox}

\begin{tcolorbox}[
    colback=orange!5,
    colframe=orange!40,
    arc=2mm,
    boxrule=0.8pt,
    left=2mm,
    right=2mm,
    top=2mm,
    bottom=2mm,
    title=\textbf{Guarded Prompt for the Summarization task}
]
\small

You are a clinical handoff assistant helping prepare a concise clinician-to-clinician summary for the next provider taking over care.

Your task is to summarize the source text into a brief clinician-facing handoff summary.

The output will be read by another clinician during a care transition. Focus on the most clinically important information needed for continuity of care, including major problems, key findings, current status, and important near-term considerations. Keep the summary concise, accurate, and professionally written.

Use only information that appears in the source text. Do not add new facts, interpretations, diagnoses, actions, recommendations, timelines, or causes. Do not convert uncertain or hedged statements into definite statements. Preserve the original meaning and level of detail as closely as possible. Improve clarity, grammar, fluency, and organization. Keep the tone professional and neutral. Output only the rewritten text.

\{\{input text\}\}
\end{tcolorbox}

\begin{tcolorbox}[
    colback=orange!5,
    colframe=orange!40,
    arc=2mm,
    boxrule=0.8pt,
    left=2mm,
    right=2mm,
    top=2mm,
    bottom=2mm,
    title=\textbf{Guarded Prompt for the Revision task}
]
\small

You are a patient education assistant helping rewrite clinical text into plain-language information that a patient or family member can understand.

Your task is to rewrite the source text in clear, patient-friendly language without changing its meaning.

The output will be read by a patient or caregiver. Use plain, direct, non-technical language where possible. Preserve the clinical meaning, but make the wording easier to understand. Keep the tone respectful, calm, and supportive.

Use only information that appears in the source text. Do not add new facts, interpretations, diagnoses, actions, recommendations, timelines, or causes. Do not convert uncertain or hedged statements into definite statements. Preserve the original meaning and level of detail as closely as possible. Improve clarity, grammar, fluency, and organization. Keep the tone professional and neutral. Output only the rewritten text.

\{\{input text\}\}
\end{tcolorbox}

\section{Prompts in RQ2 Experiments}\label{appendix:rq2_prompt}

\begin{tcolorbox}[
    colback=blue!5,
    colframe=blue!40,
    arc=2mm,
    boxrule=0.8pt,
    left=2mm,
    right=2mm,
    top=2mm,
    bottom=2mm,
    title=\textbf{Direct Uncertainty Classification Prompt}
]
\small

You will read a short clinical text snippet. Focus only on the target concept given below. Ignore uncertainty language that applies to other concepts in the snippet.

Uncertainty level definitions: 

UL1: the concept is explicitly absent or ruled out; this is a hard negative. 

UL2: the author favors the concept, but does not assert it definitively. 

UL3: the concept is plausible, but the author shows weaker commitment than UL2. 

UL4: the text explicitly says the concept is unresolved, indeterminate, or cannot currently be decided. 

UL5: the concept is mentioned as a query, rule-out target, workup target, follow-up target, or patient-specific future contingency, rather than as a current conclusion about the patient's present state.

Task: 1. Find the shortest quote in the snippet that signals the uncertainty status of the target concept. 2. Assign the correct uncertainty level for that target.

Output rules: Return JSON only. Do not wrap the JSON in markdown code fences. Do not include any explanation outside the JSON. Copy the evidence quote exactly from the snippet. Judge only the target concept, not the whole sentence. If several phrases are present, choose the one that most directly determines the uncertainty level for the target. Quote the uncertainty signal, not the target phrase itself. Use only the text in the snippet. Do not use outside clinical knowledge. Use only one of these labels: UL1, UL2, UL3, UL4, UL5. Use UL5 narrowly: if the text expresses a current possibility or probability about the target, choose UL2 or UL3 instead of UL5. If the text says the target cannot currently be resolved, choose UL4 instead of UL5. If the best evidence is punctuation such as ``?'', copy that exact symbol as the evidence quote.

Output schema:
\{\{``target\_id'': ``...'', ``target\_text'': ``...'', ``evidence\_quote'': ``...'', ``uncertainty\_level'': ``UL1 or UL2 or UL3 or UL4 or UL5''\}\}

Target:
target\_id: \{target\_id\}, target\_text: \{target\_text\}

\{\{input text\}\}
\end{tcolorbox}

\begin{tcolorbox}[
    colback=blue!5,
    colframe=blue!40,
    arc=2mm,
    boxrule=0.8pt,
    left=2mm,
    right=2mm,
    top=2mm,
    bottom=2mm,
    title=\textbf{Multi-Signal Uncertainty Classification and Ranking Prompt}
]
\small

You will read multiple short clinical text signals. Each signal contains one target concept and one short snippet. Your job has two parts.

For each signal, find the shortest quote in the snippet that determines the uncertainty status of the target, and assign the correct uncertainty level.

After you have labeled all signals, rank them from strongest to weakest author commitment about the target as a current patient-state claim.

Uncertainty level definitions: 

UL1: the concept is explicitly absent or ruled out; this is a hard negative. 

UL2: the author favors the concept, but does not assert it definitively. 

UL3: the concept is plausible, but the author shows weaker commitment than UL2. 

UL4: the text explicitly says the concept is unresolved, indeterminate, or cannot currently be decided. 

UL5: the concept is mentioned as a query, rule-out target, workup target, follow-up target, or patient-specific future contingency, rather than as a current conclusion about the patient's present state.

Ranking order: For ranking, use this fixed order of author commitment, strongest to weakest: UL1 $>$ UL2 $>$ UL3 $>$ UL4 $>$ UL5. Rank strength of author commitment about the patient's current state. Do not rank disease severity, clinical importance, or treatment priority. Judge each signal separately before ranking the full set. Use only the text in each snippet. Do not use outside clinical knowledge. Focus only on the anchored target for each signal. Ignore uncertainty language that applies to other concepts in the same snippet.

Output rules: Return JSON only. Do not wrap the JSON in markdown code fences. Do not include any explanation outside the JSON. Copy each evidence quote exactly from the corresponding snippet. Use the shortest quote that most directly determines the uncertainty level. Use only one of these labels: UL1, UL2, UL3, UL4, UL5. Use UL5 narrowly: if the text expresses a current possibility or probability about the target, choose UL2 or UL3 instead of UL5. If the text says the target cannot currently be resolved, choose UL4 instead of UL5. If the best evidence is punctuation such as ``?'', copy that exact symbol as the evidence quote. The signals list in your output must contain every input signal exactly once. The rank\_strongest\_to\_weakest list must contain every signal\_id exactly once, with no duplicates and no omissions.

Output schema: \{\{``signals'': [\{\{``signal\_id'': ``S1'', ``target\_text'': ``...'', ``evidence\_quote'': ``...'', ``uncertainty\_level'': ``UL1 or UL2 or UL3 or UL4 or UL5''\}\}], ``rank\_strongest\_to\_weakest'': [``S1'', ``S2'', ``S3'']\}\}

\{\{input text\}\}
\end{tcolorbox}

\section{BERTScore for RQ1 Outputs}\label{appendix:bertscore}

\begin{table*}[htbp!]
  \centering
  \caption{BERTScore-F1 between source documents and LLM-generated outputs in RQ1. All values are percentages. Despite high semantic similarity across all conditions, LLMs do not preserve uncertainty cues correctly (see Section~\ref{subsec:rq1_results}).}
  \small
  \begin{tabular}{llc}
    \toprule
    \textbf{Model} & \textbf{Task / Condition} & \textbf{BERTScore F1} \\
    \midrule
    \multirow{4}{*}{\texttt{claude-haiku-4.5}}
      & Revision / Baseline       & 82.00 \\
      & Revision / Guarded        & 83.67 \\
      & Summarization / Baseline  & 83.45 \\
      & Summarization / Guarded   & 85.93 \\
    \midrule
    \multirow{4}{*}{\texttt{gemini-2.5-flash}}
      & Revision / Baseline       & 82.20 \\
      & Revision / Guarded        & 83.94 \\
      & Summarization / Baseline  & 84.65 \\
      & Summarization / Guarded   & 86.46 \\
    \midrule
    \multirow{4}{*}{\texttt{gpt-oss-120b}}
      & Revision / Baseline       & 82.06 \\
      & Revision / Guarded        & 84.04 \\
      & Summarization / Baseline  & 82.34 \\
      & Summarization / Guarded   & 84.64 \\
    \bottomrule
  \end{tabular}
  \label{tab:bertscore_results}
\end{table*}